\documentclass[runningheads,a4paper]{llncs}

\usepackage{amssymb}
\usepackage{graphicx,epstopdf}

\usepackage{url}
\urldef{\mailsa}\path|{alfred.hofmann, ursula.barth, ingrid.haas, frank.holzwarth,|
\urldef{\mailsb}\path|anna.kramer, leonie.kunz, christine.reiss, nicole.sator,|
\urldef{\mailsc}\path|erika.siebert-cole, peter.strasser, lncs}@springer.com|
\newcommand{\keywords}[1]{\par\addvspace\baselineskip
\noindent\keywordname\enspace\ignorespaces#1}

\begin{document}

\mainmatter  

\title{Face Recognition: Perspectives from the Real-World}

\titlerunning{Face Recognition: Perspectives from the Real-World}

%
%
\author{Bappaditya Mandal}
\authorrunning{Bappaditya Mandal}

\institute{Institute for Infocomm Research, A*STAR, \\ 1 Fusionopolis Way, \#21-01 Connexis (South Tower), Singapore 138632. \\ Phone: +65 6408 2071; Fax: +65 6776 1378; \\ E-mail: bmandal@i2r.a-star.edu.sg}

%
%

\toctitle{Lecture Notes in Computer Science}
\tocauthor{Authors' Instructions}
\maketitle

\begin{abstract}
In this paper, we analyze some of our real-world deployment of face recognition (FR) systems for various applications and discuss the gaps between expectations of the user and what the system can deliver. We evaluate some of our proposed algorithms with ad-hoc modifications for applications such as FR on wearable devices (like Google Glass), monitoring of elderly people in senior citizens centers, FR of children in child care centers and face matching between a scanned IC/passport face image and a few live webcam images for automatic hotel/resort checkouts. We describe each of these applications, the challenges involved and proposed solutions. Since FR is intuitive in nature and we human beings use it for interactions with the outside world, people have high expectations of its performance in real-world scenarios. However, we analyze and discuss here that it is not the case, machine recognition of faces for each of these applications poses unique challenges and demands specific research components so as to adapt in the actual sites.
\end{abstract}

\section{Introduction}
\label{sec:intro}

Face recognition (FR) has been a very demanding and an important core functional component in many vision based systems performing large number of social, corporate and commercial activities \cite{Mandal2}. They involve important applications in security, law enforcement, health-care, entertainment, e-commerce and so on and so forth \cite{TedNews1}. It has advantages over other biometric technologies (such as finger print, ear, palm print, iris, etc) because it is non-intrusive, non-invasive and easy-to-use. It is a technology that can be used to identify an individual at a distance without the cooperation or knowledge of the subject. It is expected that a FR system should automatically identify and verify individuals in videos/still images. The former compares a query face image against all the stored templates in the database to determine the identity of the query face (one-to-many matching) \cite{Jiang3}. While the later compares a query face image of supposedly known person against this person's claimed identity stored in the system (one-to-one matching). The third scenario involves a watch-list check, where a query face is matched to a list of suspects (one-to-few matching).

The recent terrorist attack in Paris that killed 129 people \cite{CNNnews1} and past many such incidents reiterate the necessity of FR technology that can automatically segment potentially dangerous people and even create alerts in the suspected areas/regions. A recent case study on unconstrained FR using the Boston marathon bombings suspects \cite{MSU1} reveals that with a commercial face matcher, the traditional challenges of FR such as pose, occlusions and resolutions continue to be the biggest hurdles in achieving rank-one hit for suspect Dzhokhar Tsarnaev against one million mugshot background database \cite{MSU1}. Even after four decades of intensive research in machine FR, the problem is still far from solved for large scale unconstraint FR.

Recent advances of FR in unconstrained framework using deep learning, like the ones reported on YouTube and `labeled faces in the wild' (LFW) databases in \cite{Taigman1,Yisun1} have shown significant performance improvement over the state-of-art FR algorithms in Face Recognition Vendor Test (FRVT 2014) \cite{FRVT2013}. However, more recently a new benchmark protocol, called BLUFR on large scale unconstrained FR is reported in \cite{Liao2}. Unlike many previous studies on LFW, BLUFR uses the entire LFW database for creating large number of genuine matches ($\approx$156,915) and imposter matches ($\approx$46,960,863). This large scale FR algorithm evaluation with new protocol reveals that only 41.66\% verification rates can be obtained at 0.1\% false acceptance rates (FAR) and 18.07\% open-set identification rates at rank 1 and 1\% FAR \cite{Liao2}. Hence, these numbers depict that FR problem is still largely unsolved and further attention and efforts are required to develop new feature representations and learning algorithms that can advance the algorithm development for FR.

In the following sections we take up some of the case studies and discuss the adaptation or new methodologies that we proposed to overcome the problems in FR from the prospectives of the real-world implementations. Specifically, we discuss the challenges, problems faced and general performance analysis for FR on wearable devices, monitoring of elderly people in senior care centers, recognition of children in child care centers and face matching in automatic checkouts in hotels/resorts.

\section{Face Recognition on Wearable Devices}
\label{sec:format}
One of our primary goals is to develop FR system on wearable devices (like GG \cite{GoogleGlass1} and GoPro \cite{GoPro1}), which will assist people in knowing identity or become familiar (seen/not seen) with an individual \cite{Mandal8,Gan1} in general meeting or business networking sessions (like attending a conference) or assist people with cognitive disability. It serves as a visual memory aid for people with visual impairment or those who cannot remember or recognize human faces (prosopagnosia \cite{Edward1}) for first-person-views (FPV) or egocentric views \cite{Gan2}.

\subsection{System Overview: Face, Eye Detections and Recognition}
\label{sec:Overview}
\begin{figure}[!htp]
\centering
\includegraphics*[width=4.8in]{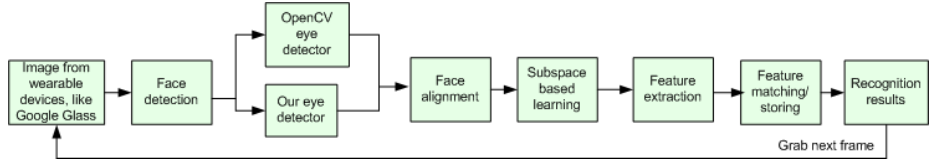}
\caption{System diagram} \vspace{-0.2cm}
\label{fig1SystemDiagram}
\end{figure}
Our system framework is shown in Fig. \ref{fig1SystemDiagram}. In the incoming videos/images we perform face and eye  detections. OpenCV face and eye detections \cite{OpenCV1} are used to localize the face and eyes. However, they cannot detect face/eye with pose variations. In a ten-minute video recording of an interaction involving 3 people, the OpenCV face detector can only detect about 40\% of the faces in the poses of looking at the GG (frontal faces). At all other times it fails due to non-frontal view faces, motion blurry, severe shadows and other poor lighting conditions. Therefore, we train a new face detector specially for the non-frontal view faces using Harr features and Adaboost classifier. Both face detectors are applied to find faces in the input image frame. We use a fusion of re-trained OpenCV eye detector \cite{OpenCV1} and our developed ISG (Integration of Sketch and Graph patterns) \cite{Yu2} eye detectors for detecting and locating a pair of open and closed eyes in both frontal-view and oblique-view faces. A systematic evaluation of FPV eye detection performance is reported in \cite{Mandal8}.

There are some unique challenges of performing FR on FPV data generated from wearable devices. One of them is that in FPV videos, both wearable camera and the subject are moving or jittering, so the images are often blurry in nature and mug shot (studio or controlled condition) image of the person is not readily available/possible. Also, it is difficult to obtain large number of images of the person to be recognized because the person might not stay in the view for a long time. Moreover, in wearable devices the computation resources are limited so the algorithm should be fast enough to be executed under constrained mobile environment \cite{Mandal6,Mandal1}.

In this system, the training face images of each person are clustered into subclasses using spatial partitioning trees and then we compute the within-subclass scatter matrix. Eigenfeature regularization (ERE) scheme \cite{Jiang4,Mandal5} is applied to regularize features obtained from whole space within-subclass scatter matrix. On these regularized features, total-subclass and between-subclass scatter matrices are computed \cite{Mandal3,Mandal8}. Low dimensional discriminative features are extracted after the whole space subclass discriminant analysis (WSSDA) \cite{Mandal12}. Cosine distance measure with first nearest neighborhood classifier (1-NN) are applied to evaluate our system. WSSDA which is based on ERE is reported to be the best performer among many state-of-the-art methodologies \cite{Mandal12} on the popular unconstrained YouTube face database \cite{Wolf1}. Hence we use these approaches or their modified versions for our evaluations.

\subsection{Performance of the System}
\label{sec:Performance}
\begin{figure}
\centering
\includegraphics[width=60mm]{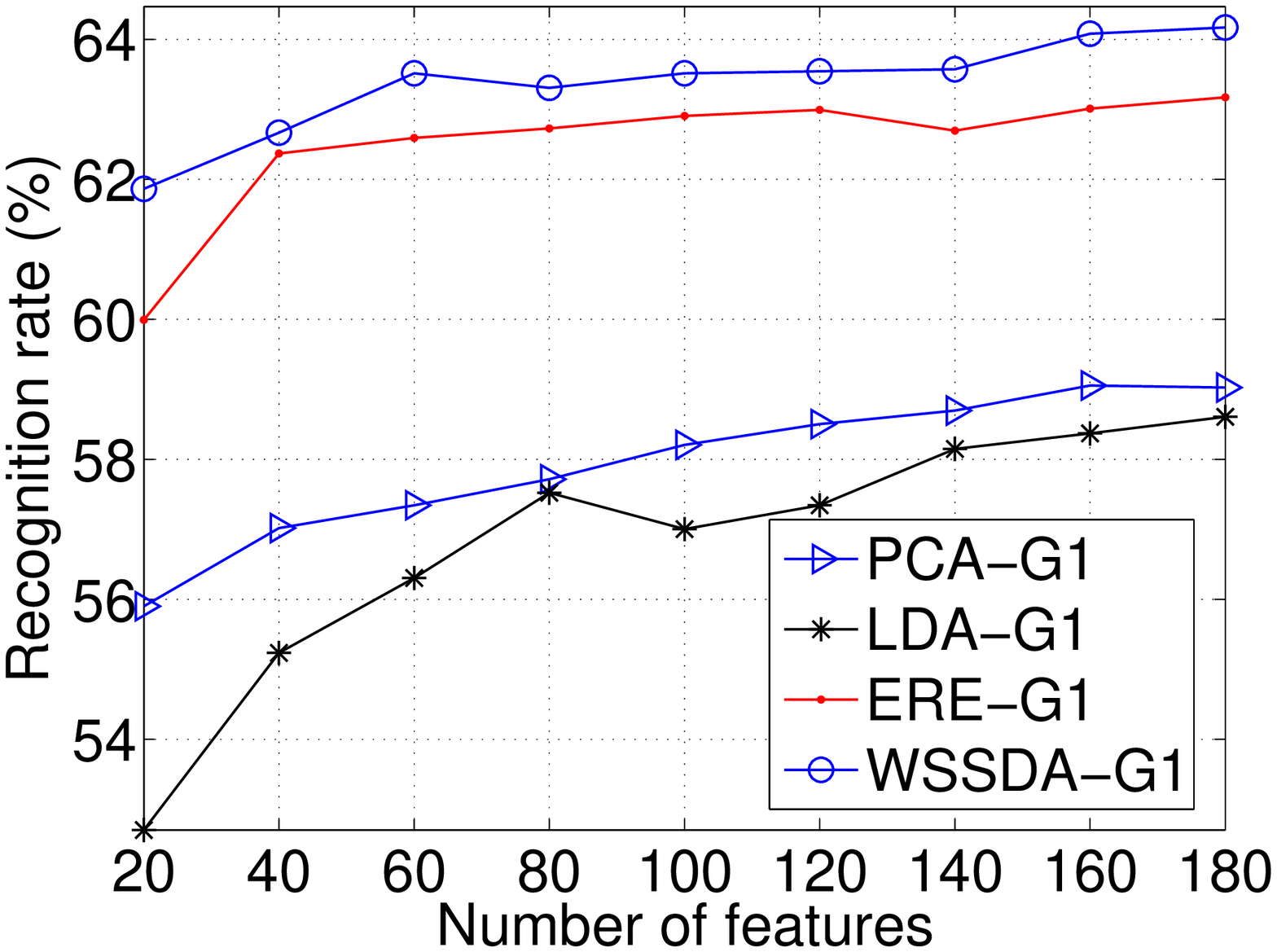}
\includegraphics[width=60mm]{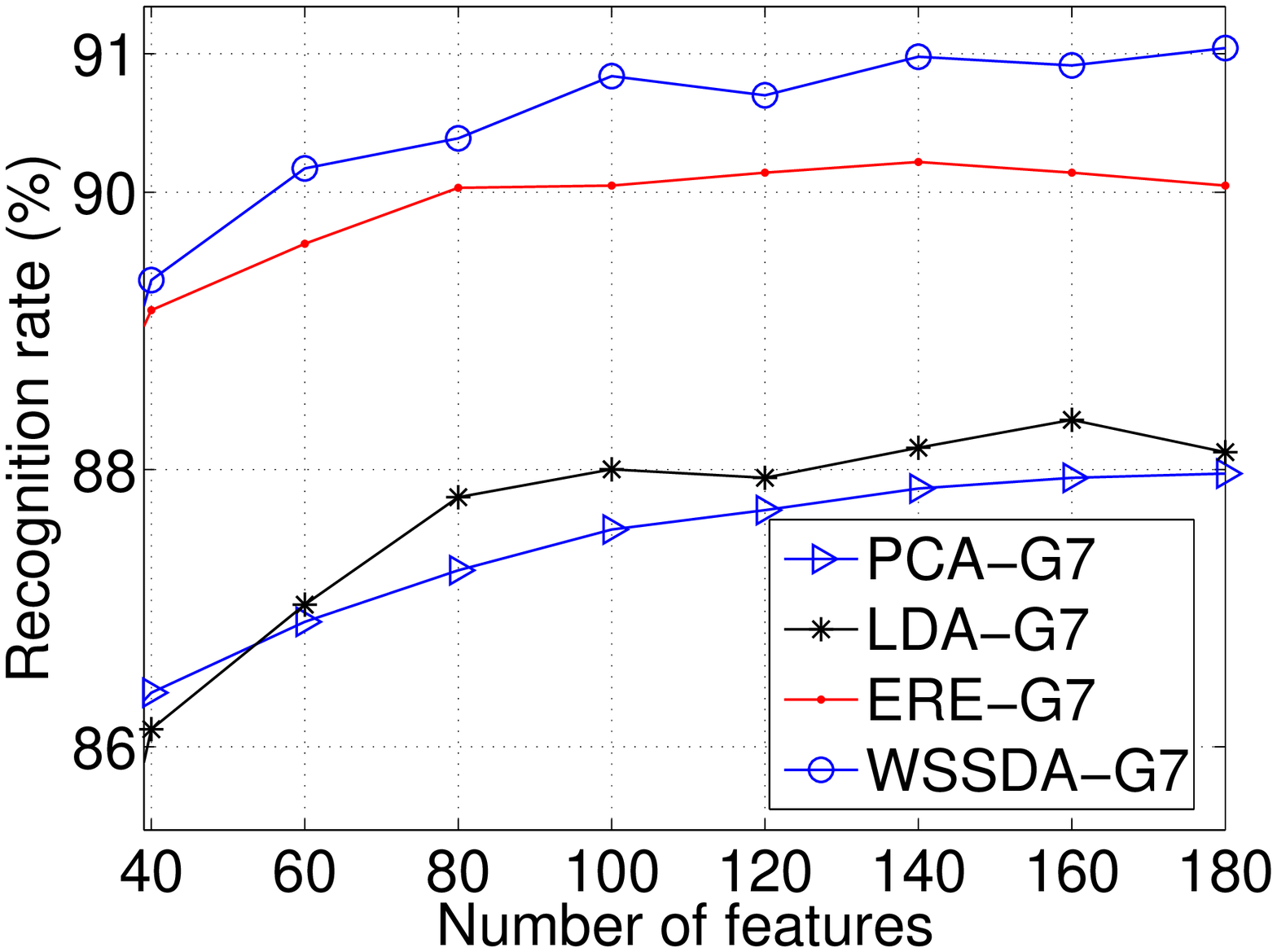}
\caption{Recognition rate vs Number of features used in the matching on wearable device database for two scenarios: Left (G1), 1 image per person in the gallery, rest all as probe images. Right (G7), 7 images per person in the gallery, rest all as probe images (best viewed in color).}
\label{ERErecoVSnoOfFeaturesGG}
\end{figure}

We evaluate the proposed WSSDA algorithm on the database built from FPV videos comprising of 7075 images of 88 subjects (average 80.4 images per subject) for FR \cite{Mandal8}. We randomly select images of 42 people for training and images of the remaining 46 people are used for testing. We test the performance of face recognition for two application scenarios. In the first case, only one frontal view face image for each person is stored in the gallery database, rest all images in the probe database (termed as G1 for each of the compared methods in Fig.~\ref{ERErecoVSnoOfFeaturesGG}). This is similar to the commercial database of personal information containing only one mug shot image for each person. In the second case, we select 7 images of different poses for each person and use them to form the gallery database and the remaining images are used as probe images (termed as G7 for each of the compared methods in Fig. \ref{ERErecoVSnoOfFeaturesGG}). Two methods, {\em i.e.} linear discriminant analysis (LDA) and principal component analysis (PCA) are selected as baseline methods for comparison as they are recommended for wearable FR in \cite{Krishna1}. The comparison with previous method of eigenfeature regularization and extraction (ERE) on within-class subspace \cite{Jiang4} is also performed.

Fig.~\ref{ERErecoVSnoOfFeaturesGG} shows the plots of recognition rates (\%) against the number of features used in the matching for two application scenarios: G1 (left) and G7 (right). It can be seen that the first scheme with only one mug shot image per person in the gallery database cannot generate satisfied result for wearable FR. In the second scheme, when using dimension of 100 features, the accuracy rate is close to 91\%. For any kind of wearable device like GG, it is really very important to achieve good recognition rates while using small number of features. Evidently, a little more efforts would be required to build the gallery database as compared to the first simpler scheme in real-world applications.

\section{Monitoring of Elderly People}
\label{sec:MonitoringElderly}
\subsection{Background}
Senior citizens in Singapore above 65 years old will grow to more than 900,000 out of a population of around 5.3 million by 2030 \cite{MSFwebsite1}. With aging population in Singapore more and more elderly people would need care giving centers for performing their daily activities \cite{Mandal7}. At these centers, elderly people engage themselves in social interactive activities such as regular exercises, play games, celebrate birthdays and festivals. So the elderly people would walk-in these centers and stay from 9am to 5pm. The authority or the care-givers faces four main problems:
(i) \textit{Alert when someone stays too long in the toilet} (may be because of a fall).
(ii) \textit{Search for the location of a particular person} (for numerous reasons, like relative wants to meet him/her).
(iii) \textit{Anybody left out in room} (like in case of  emergency).
(iv) \textit{Find anyone left the center without informing the care givers.}
We plant a set of IP cameras in the elderly care center to capture face images of the elderly people to recognize each individual in a non-intrusive way along with the hourly time sequence of the day and try to achieve the above functionalities.

\subsection{System Setup}
Our deployment involves a network of 11 IP cameras placed at strategic locations in an Elderly care center in Singapore \cite{MoMwebsite1} so as to obtain the above four functionalities. The cameras are connected to the three desktop PCs placed in the centralized computation room inside the elderly care center. All these functionalities require robust algorithms for (i) human detection, (ii) face detection, (iii) eye detection and (iv) face recognition. The cameras for FR are installed at exit and/or entrance, path ways to washroom and other strategic areas at a level so as to capture frontal view of face images. The unique challenges of these face images obtained from the elderly people are:
(i) Elderly people, due to aging tend to bend forward posing more challenges in capturing good frontal face images.
(ii) Faces of elderly people have large wrinkles and other artifacts. During normalization many more noises are introduced into the templates or the matching images \cite{Mandal9}.

Human detection \cite{Satpathy1} is applied to find the region of interests in the incoming images from the IP cameras. Face detection \cite{Viola04} re-trained with elderly people is used to get the face coordinates. On the detected faces, eye detections \cite{Yu2,OpenCV1} are applied. Using the eye coordinates, faces are cropped and normalized following \cite{Beveridge}. Our proposed eigenfeature regularization and extraction (ERE) approach \cite{Mandal3} is used to train and evaluate this elderly care monitoring system.

\subsection{Performance Evaluation}
\begin{figure}
\centering
\includegraphics[width=60mm]{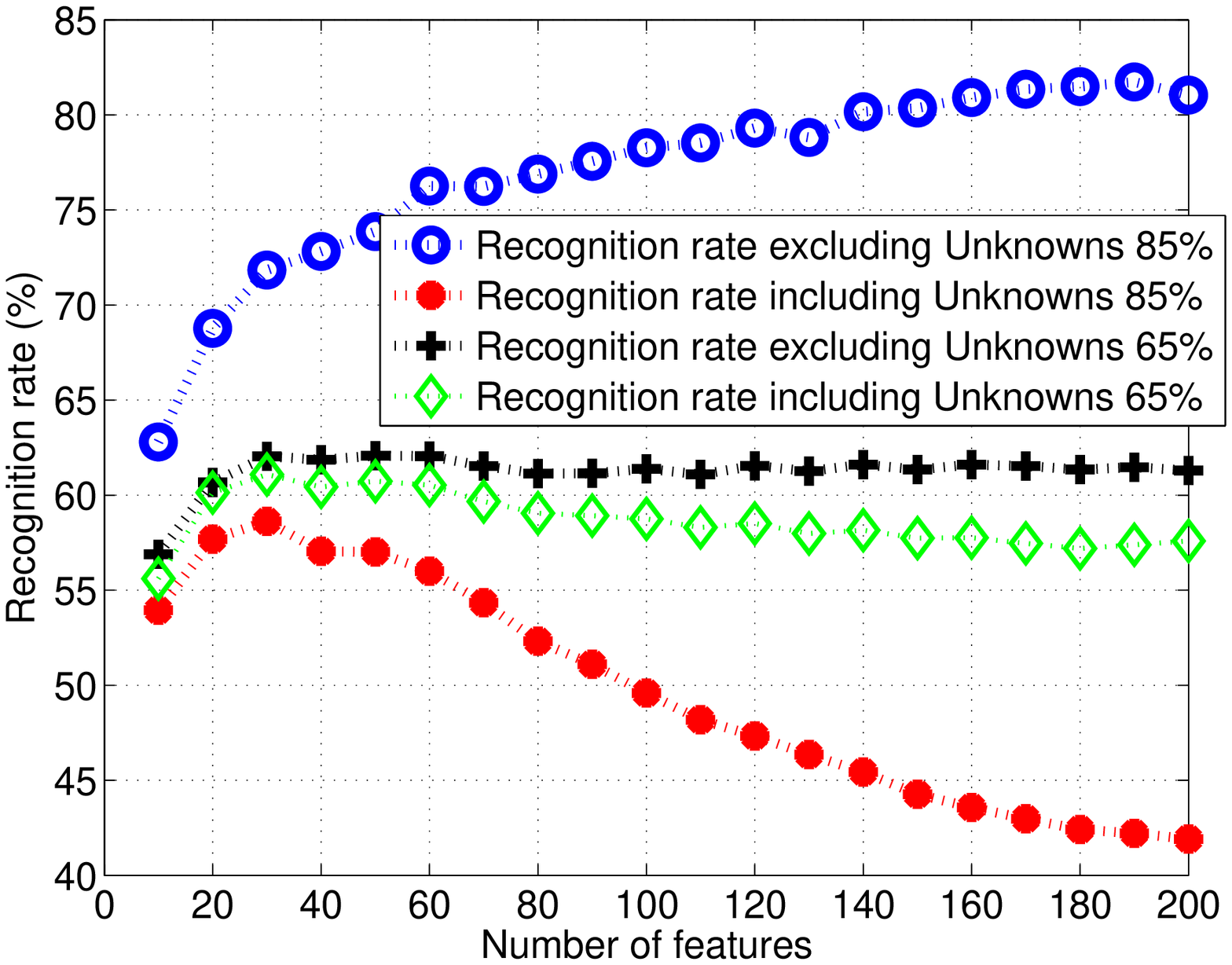}
\includegraphics[width=60mm]{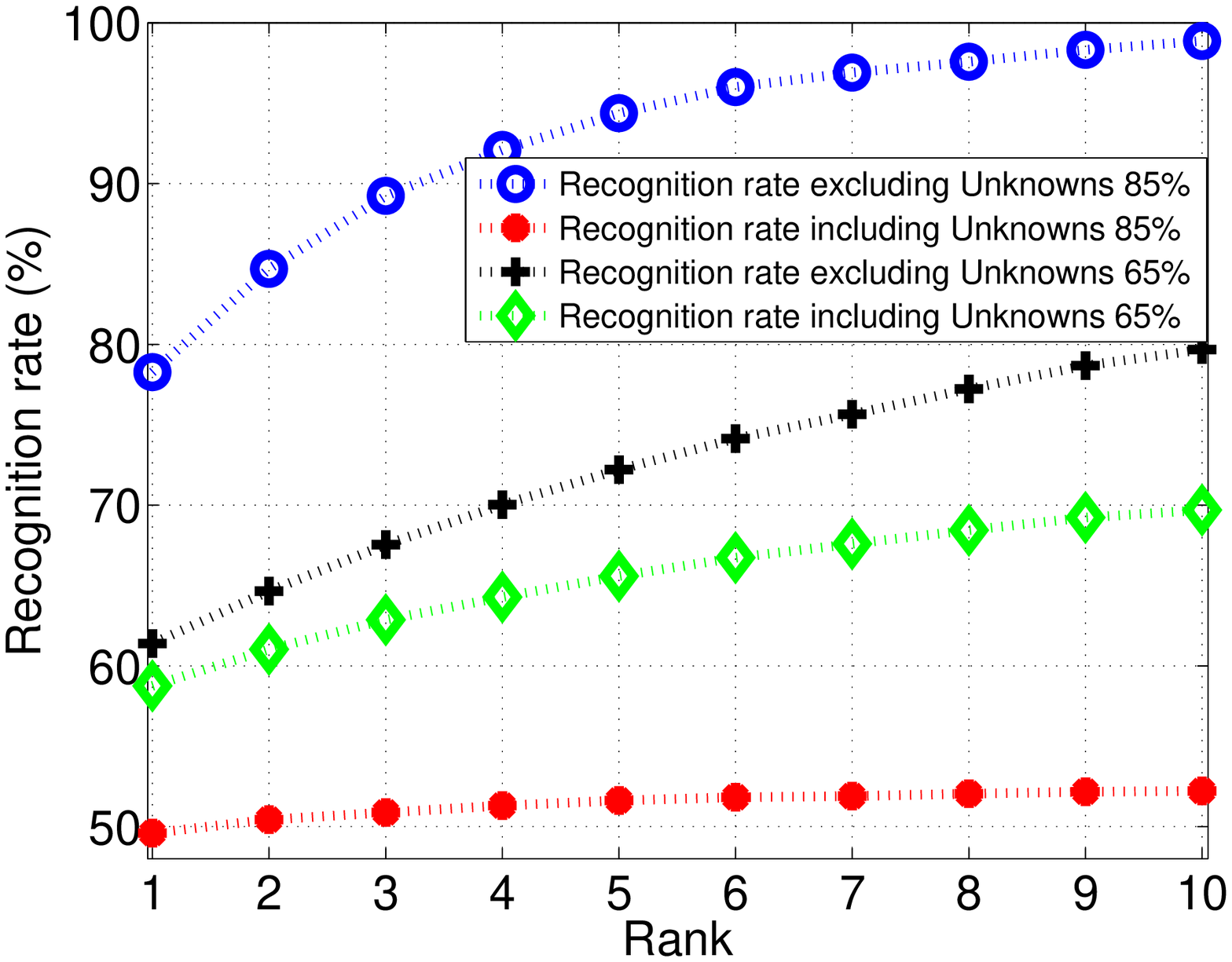}
\caption{Left, Face identification rate (\%) vs. number of features used in the matching. Right, Face identification rate (\%) vs. rank with 70 features (best viewed in color).}
\label{ERErecoVSnoOfFeaturesRankETH}
\end{figure}

We collected about 40 hours of active face images of the elderly people and care-givers. From this videos, 60 elderly people with total of 4450 face images are used for testing. The longest and shortest number of images per person is 465 and 9 frames, respectively, with an average of 74.2 frames per person.  These elderly people are registered (or the system is trained) with 7 images per person, each using images obtained from similar IP cameras. We test our system on two scenarios: face identification (FI) as the top 1 match with cosine distance measure against the number of features used in the matching and FI along with rank plots with 70 features as shown in Fig. \ref{ERErecoVSnoOfFeaturesRankETH}.

We set the threshold as 85\% of the max distance measure value among all the registered people (genuine probes). Any distance measure less than this threshold would be used for identification, rest all are taken as `Unknowns' (imposter probes). Similarly we test for 65\% of the max distance measure value. Fig. \ref{ERErecoVSnoOfFeaturesRankETH} shows the performance evaluations on the data collected in one of the elderly care centers. Excluding the `Unknowns', left of Fig. \ref{ERErecoVSnoOfFeaturesRankETH}, we could achieve 80\% accuracy rate. Whereas for care givers, if they query any person and retrieve top 10 matches with 70 features, right of Fig. \ref{ERErecoVSnoOfFeaturesRankETH}, then 97\% accuracy can be obtained. Including `Unknowns' as imposters, these numbers are 63\% and 78\% respectively.

\section{Recognizing Children in Child Care Centers}
\label{sec:childcareCenters}
\subsection{Background}
One of the important tasks in Singapore child care centers is the daily body temperature measurement and tagging them to each of the individual's identity. Every morning when the child comes to a child care center, its body temperature is taken so as to get some idea of any fever or hand-to-mouth disease or early prevention of infectious diseases (or epidemic like SARS \cite{SARSwebsite1}). The body temperature measurement is crucial for many infectious diseases. Once a child detected with high temperature, special care or monitoring of the child is performed by the care givers in child care centers.

\subsection{System Configuration}
The entry of the child care center is equipped with a long vertical slim standing bar, slightly longer than the longest possible child in the care center. This bar has a thermal cameras which senses the body temperature of the child non-intrusively. Once the temperature is taken, the care giver would manually tag its identity (name) to this temperature for that day. This is done for every working day, which is labor intensive and raises the cost of child care centers. So we set ourselves a task where after taking the body temperature, this bar would have a camera, which would automatically find the identity of the child and tag its temperature automatically. So no manual intervention would be required and it will reduce the burden on the care givers. We process and recognize the children face images using the similar framework as described in Secs. \ref{sec:Overview} and \ref{sec:Performance}.

\subsection{Performance Evaluation}
Our efforts for child care centers would help in automatic tagging of body temperature to the identity (`without touch sensors') of the children. The challenges involved in such deployment are: (i) children faces are small, so their distinctness is reduced to a large extend, (ii) they are uncooperative and whimsical in nature, so getting them to pose for frontal or near frontal images is extremely difficult, (iii) as a result of which, their registration is difficult and has to be repeated again and again, finally, (iv) their faces change very fast because of their growing nature as compared to average adults. We obtained a database of 14 children comprising of 119 images, average of 8.5 images per child. Similar to the above performances, we could obtain an average recognition rate of 64\% using only 80 features.

\section{Face Matching for Automatic Checkout}
\subsection{Background}
\begin{figure}
\centering
\includegraphics[width=120mm]{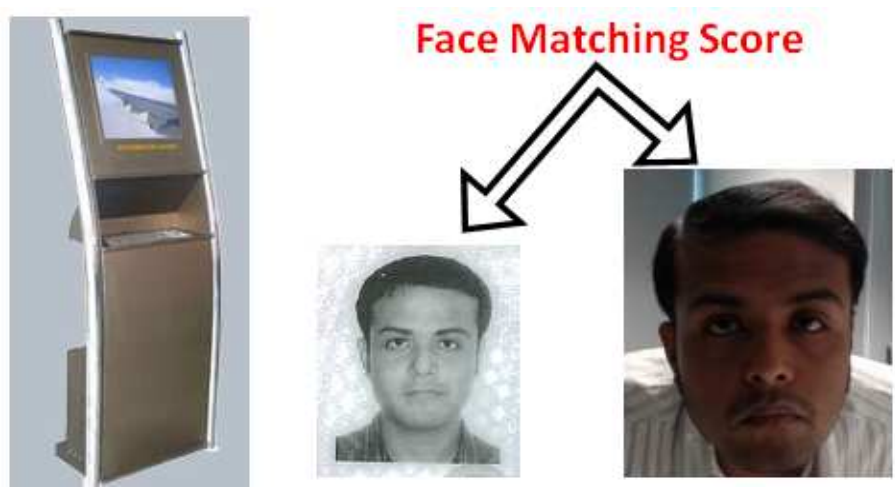}
\caption{Left: A Kiosk, Middle: A sample face from the scanned IC and Right: A live webcam face image (best viewed in color).}
\label{Kiosk}
\end{figure}

We describe here a practical application where a kiosk is developed for automatic checkout of hotel/resort guests. The system has a scanning device and a webcam connected to a centralized server. When a person is checking out from the hotel, he/she is expected to present his IC/Passport onto a kiosk as shown in Fig. \ref{Kiosk}, left. The system captures his passport image and performs face detection so as to localize the face. The webcam from the kiosk, in front of the person also captures his live face images. Due to the local police regulations and time constraints for the user, it can capture and use only a few face images. These face images are matched against the single passport face image and a few matching scores are generated. Based on these matching scores the system tries to verify whether the passport face and the live captured face images are same or different.

The challenges involved in such face matching process is that there is no scope for training. Choosing appropriate threshold is very critical and difficult for developing such systems. The system has to generate reasonably well and judiciously chosen threshold so as it can infer whether the (scanned and live) face images are same or different.

The greatest challenge in this problem is the age difference between the IC or passport face image and the live face images, as shown in Fig. \ref{Kiosk}, middle and right. This is because often the IC or passport images are captured about 5-8 years ago (from recent to 10 years), the face changes considerably over this period of time and hence, the faces are very difficult to match. Moreover, the IC or passport face images are often watermarked or mosaiced, their quality are poor and noisy. After they are scanned, these images become more noisy and blurry. When matched against the live webcam images the matching scores are highly unpredictable owing to the complex pattern of the face images. Although, all human beings has two eyes, one nose, one mouth, forehead and the cheek regions, the distinguishable characteristics of one person from another are the key factors that the machine learning system needs to consider \cite{Mandal13}. We have tried to develop such system in the real-world that can match a scanned face image with that of a few live face images captured using a webcam.

\subsection{Performance Evaluation}
Our system is pre-trained with 88 person comprising of 7075 images as described in \cite{Mandal8}. We preprocess and normalize all the face images using the normalization technique described in \cite{Mandal3}. On a small test database of 33 images with 8 subjects' images, we could achieve about 50\% accuracy in the real-world settings. So all the 4 subjects are recognized correctly using their IC/passport images and live face images above a single specific threshold.

\section{Summary and Conclusions}
\label{sec:summary}
Out of many deployments, we describe 4 cases where FR systems are deployed and test-bedding are performed. They are applied to obtain various functionalities and make the existing systems/protocol more robust, efficient, less manual intervention and increase productivity. They pose unique challenges and difficulties in actual sites and also to the computational models/algorithms development. We have collected real-world face image data for each of these applications and applied our previously proposed popular approaches like ERE and WSSDA. These algorithms are modified and evaluated for all these applications. The system performances that we have obtained in all cases are comparatively low as compared to what the users would expect. In other words, these evaluations on real-world FR system deployments show that machine recognition of humans are still far away from the human recognition of faces. Our evaluations show that the real deployment of FR in actual sites require more in-depth research in specific cases and application domain.

\bibliographystyle{splncs03}
\bibliography{face}
\end{document}